\documentclass[letterpaper]{article}
\usepackage{aaai}
\usepackage{times}
\usepackage{helvet}
\usepackage{courier}
\usepackage{graphicx}
\usepackage{amsmath,amsthm,amscd,amssymb}
\usepackage{subfigure}

\newcommand{\hide}[1]{}

\begin{document}

\title{The Directed Closure Process in Hybrid Social-Information Networks, \\ with an Analysis of Link Formation on Twitter}

\author{Daniel M. Romero\\
Cornell University\\
dmr239@cornell.edu\\
\And
Jon Kleinberg\\
Cornell University\\
kleinber@cs.cornell.edu\\
}

\maketitle

\newcommand{\UnnumberedFootnote}[1]{{\def\thefootnote{}\footnote{#1}
\addtocounter{footnote}{-1}}}

\UnnumberedFootnote{
This work has been supported in part by
NSF grants BCS-0537606, IIS-0705774,
IIS-0910664, 
CCF-0910940, 
a Google Research Grant,
a Yahoo!~Research Alliance Grant, and
the John D. and Catherine T. MacArthur Foundation.
}

\begin{abstract}
It has often been taken as a working assumption that directed links in 
information networks are frequently formed by ``short-cutting'' a
two-step path between the source and the destination --- a kind
of implicit ``link copying'' analogous to the process of triadic
closure in social networks.
Despite the role of this assumption in theoretical models such 
as preferential attachment, it has received very little
direct empirical investigation.
Here we develop a formalization and methodology for studying
this type of directed closure process, and we provide evidence 
for its important role in the formation of links on Twitter.
We then analyze a sequence of models designed to capture 
the structural phenomena related to directed closure that we observe 
in the Twitter data.
\end{abstract}

\section{Introduction}

Information networks, which connect Web pages
or other units of information, 
and social networks, which connect people, are
related notions, but they exhibit fundamental differences.
Two of the principal differences are based on directionality
and heterogeneity.
First, information networks are generally directed structures,
with links created by one author to point to another;
social networks, on the other hand, tend to be represented 
in most basic settings as undirected structures, 
expressing relationships that are approximately mutual.
Second, information networks tend to contain a few nodes with extremely
large numbers of incoming edges --- documents or pages that
are ``famous'' and hence widely referenced ---- while 
social networks exhibit disparities in connectivity only to a smaller extent,
since even the most gregarious people have some practical limit on the number
of genuine social ties they can form.

The link structure of the Web, and of well-defined
subsets of the Web such as the blogosphere and Wikipedia, are clear
examples of information networks; social-networking sites such as
Facebook have provided us with very large representations of 
social networks that are derived from social structure in the off-line world.
An interesting recent development has been the growth of social media
sites that increasingly interpolate between the properties of 
information networks and social networks.
The micro-blogging site Twitter is a compelling example of such an
interpolation.  
A user on Twitter is able to create links to other users whose
content he or she is interested in; this is referred to 
as {\em following} these users,
and the set of all such follower relations defines a network.
The structure of this network reflects properties both of a social network,
since it exposes underlying friendship relations among people, 
and also of an information network,
since it is directed and also contains huge concentrations of links to
specific ``celebrities'' and automated generators of news content 
that reflect fundamentally informational relations.

\begin{figure}[tp]
  \begin{center}
    \subfigure[{\em Undirected feed-forward triangle}]{
       \includegraphics[scale=0.40]{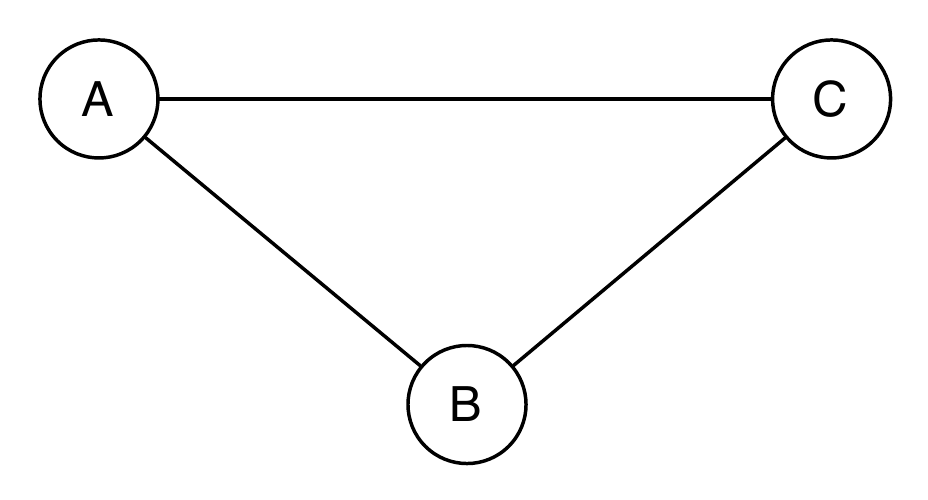}
       \label{fig:undirected-closure}
       }
    \hspace*{0.01\textwidth}
    \subfigure[{\em Directed ``feed-forward'' triangle}]{
       \includegraphics[scale=0.40]{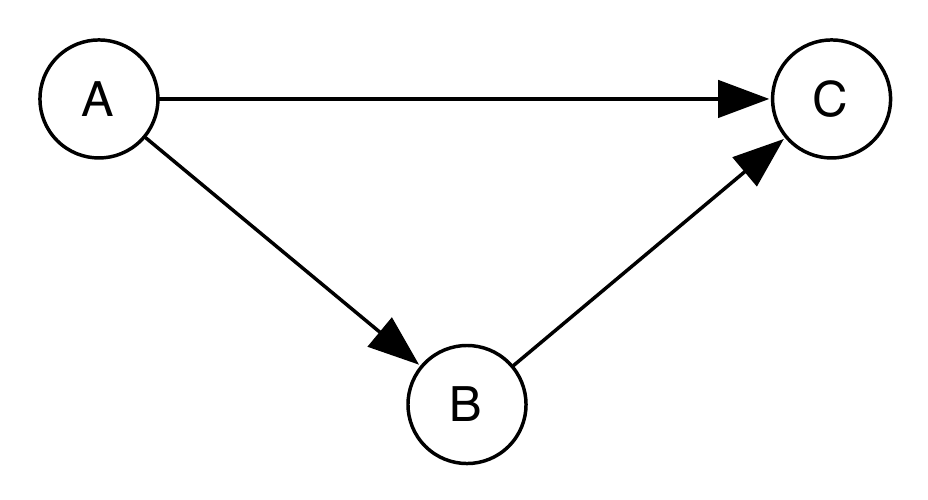}
       \label{fig:directed-closure}
       }
  \end{center}
  \caption{(a) Triadic closure in an undirected graph produces a triangle when
an edge connects two nodes who already have a common neighbor.
(b) Analogously, in a directed information network, directed closure
occurs when a node $A$ links to a node $C$ to which it already has
a two-step path (through a node $B$).  This creates a directed triangle
(a ``feed-forward'' structure on three nodes).
  \label{fig:closure}
}
\end{figure}

\paragraph{Link Formation in Information Networks.}
In a social network, 
triadic closure is one of the fundamental processes
of link formation:
there is an increased chance that
a friendship will form between two people if they already
have a friend in common 
\cite{rapoport-triadic,granovetter-weak-ties}.
(For example, we could imagine the $A$-$C$ friendship in 
Figure~\ref{fig:undirected-closure} as forming after the
existence of the $A$-$B$ and $B$-$C$ edges, and accelerated
by the existence of these two edges.)
Recent empirical analysis has quantified this effect
on large social network datasets \cite{kossinets-email}.
Is there an analogous process in information networks?

A natural hypothesis for such a process is the following:
if a node $A$ in an information network links to $B$, and $B$ links to $C$,
then one should arguably expect an increased likelihood that 
$A$ will link to $C$ --- since the author of $A$ has an increased
ability to become aware of $C$ via the two-step path through $B$.
(See Figure~\ref{fig:directed-closure}.)
We will refer to this as the {\em directed closure process}.
In addition to its intuitive appeal, this process contains an implicit
hypothesis about how links are formed in information networks ---
through the ``copying'' of a link from something you already point to ---
and such copying mechanisms form a crucial part of the motivation
for the fundamental notion of preferential attachment 
\cite{albert-revmodphys,kumar-copying,newman-sirev}.
Despite the importance of the notion, however, 
there has been remarkably little empirical analysis of the
extent to which this type of directed closure is truly at work
in real information networks, 
and of the effects it may have on network structure.

\begin{figure}[tp]
  \begin{center}
    {\includegraphics[scale=0.4]{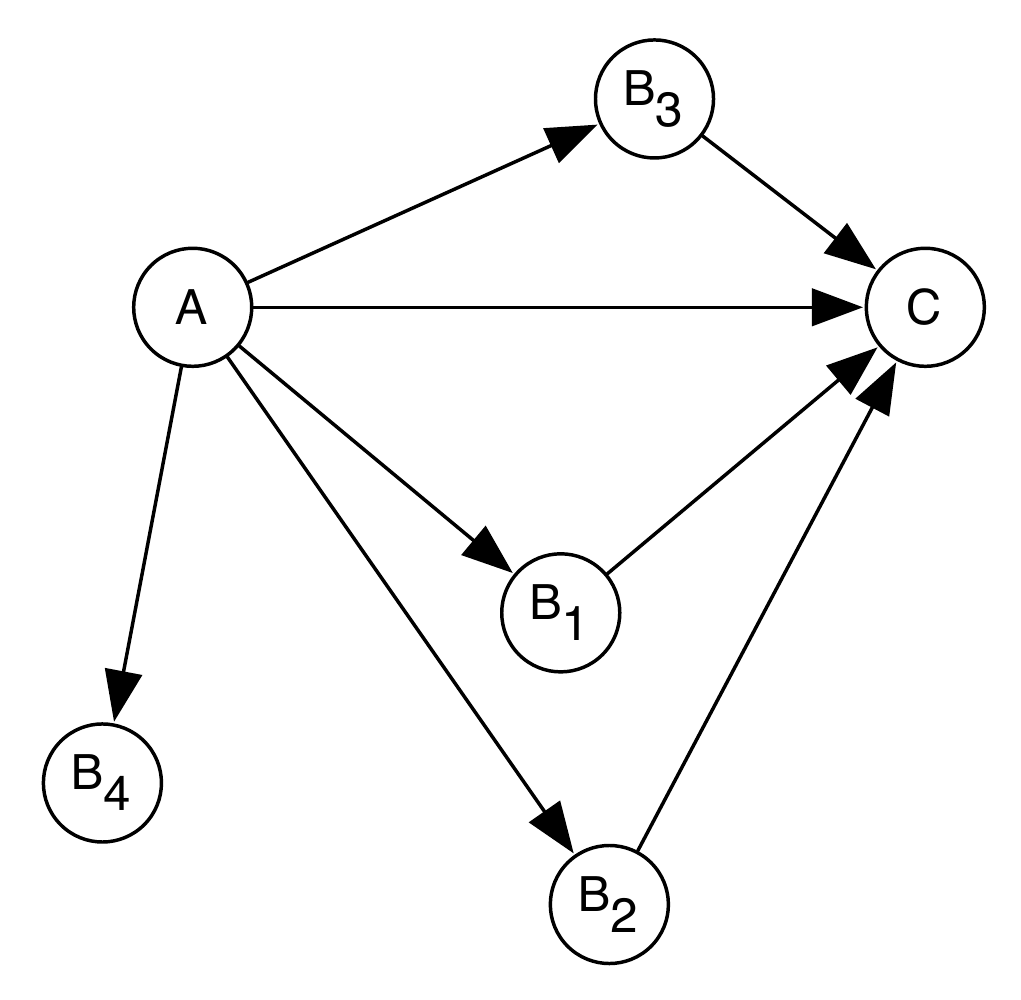}}
  \end{center}
  \caption{In this example, the edge from $A$ to $C$ exhibits closure
if there is already a two-step path from $A$ to $C$ (i.e., through 
$B_1$, $B_2$, $B_3$) when the $A$-$C$ edge arrives.
  \label{fig:multi-closure}
}
\end{figure}

\paragraph{The Present Work: The Directed Closure Process.}
In this paper, we analyze the directed closure process using data
from Twitter: we provide some of the first evidence on 
large information networks that 
directed closure is taking place at a rate significantly above 
what would be expected by chance; we identify a surprising
level of heterogeneity in how strongly it operates across
different parts of the network; and we analyze models that
capture these effects.

An important difference between triadic closure in 
social networks and directed closure in information networks
is the following observation, which in a sense serves as
the starting point for our analysis: 
while the extent of triadic closure can be assessed
from a single snapshot of an undirected graph, the
evaluation of directed closure inherently requires some form of temporal
sequence information.  
Indeed, when we see an undirected triangle such as
the one in Figure~\ref{fig:undirected-closure}, 
we know that whichever edge formed last will complete
a two-step path consisting of the earlier two edges, and hence
will satisfy the definition of triadic closure.
On the other hand, the structure in Figure~\ref{fig:directed-closure}
satisfies the definition of directed closure only if the $A$-$C$ edge
formed after the other two.

This means that the amount of directed closure in a directed graph
depends not just on the graph's structure, but also on the order in which
edges arrive.  Because of this
we are able to develop a natural {\em randomization test}
to evaluate whether directed closure is taking place in a given network
at a rate above chance.
Specifically, we say that an edge in a directed graph {\em exhibits closure}
if, at the time it forms, it completes a directed two-step path
between its endpoints.
For example, in Figure~\ref{fig:multi-closure}, the $A$-$C$ edge exhibits
closure if and only if it arrives after the pair of edges in one of the
three possible two-step $A$-$C$ paths through $B_1$, $B_2$, or $B_3$.
For a given network, we can thus ask: how many edges exhibit closure, and
how many would have exhibited closure (in expectation) if the
edges had arrived in a random order?
The point is that in any arrival order of the edges, some number of
the edges will close directed triangles; but if directed closure
is a significant effect, then we may expect to see a larger number of
such triangle-closings compared to what we'd see under a random arrival order.

To investigate this empirically, we choose a random sample of 
{\em micro-celebrities} on Twitter, which we define to be users
with between 10,000 and 50,000 followers.
(We will abbreviate the term as {\em $\mu$-celebrity}.)
For each such $\mu$-celebrity $C$, we determine the number of
edges to $C$ that exhibit closure, and compare it to the 
expected number of edges to $C$ that would exhibit closure 
in a random ordering --- we will refer to this latter number as the
{\em random-ordering baseline}.
Given that we are studying the followers of users with high numbers
of in-links, one would conjecture that there are two competing forces at work.
In one direction is the intuitively natural tendency of directed closure
to create short-cuts in the presence of two-step paths.
In the other direction, however, is the plausible tendency for 
people to link first to celebrities, before they link to more obscure users;
that is, it is not clear that closure processes are necessary
in order for people to discover and link to very prominent users.
This latter effect would tend to cause triangles as in 
Figure~\ref{fig:directed-closure} to appear with the $A$-$C$ and $B$-$C$
edges first, reducing the extent of directed closure in the real data.

We find in the Twitter data that the number of edges to a $\mu$-celebrity
that exhibit closure is higher than the random-ordering baseline,
indicating that even in linking to celebrities, there is an
above-chance tendency to do this by closing an existing two-step path.
This finding suggests a range of further interesting
questions --- specifically, whether the high rate of directed closure
is due to overt copying of follower lists (as in the intuitive basis
for the definition), or due to more subtle, implicit mechanisms 
that produce copying behavior at a macroscopic level.
To address this question, as we discuss below, 
we consider the extent to which directed
closure can arise even in models that do not explicitly build
in copying as a mechanism.

\paragraph{The Present Work: Directed Closure and Network Structure.}
Given the prevalence of directed closure in the Twitter network, 
one might suppose that it operates according to a relatively
uniform underlying mechanism.
But what we find, surprisingly, is significant heterogeneity
in the amount of directed closure.
We define the {\em closure ratio} of a $\mu$-celebrity $C$ to 
be the fraction of $C$'s incoming edges that exhibit closure.
If we track the closure ratio of $C$ as edges to $C$ are 
added in their temporal order, we find that the ratio
stabilizes to an approximately constant value fairly early.
However, the value to which the closure ratio stabilizes varies considerably
from one $\mu$-celebrity to another, and is not closely related
to the number of followers.
Thus, the closure ratio appears to be an intrinsic and diverse
property of users with large numbers of followers: some such users
receive a clear majority of their incoming links via the closing
of a directed triangle, while others receive a much smaller proportion
of their links this way.

The cause of this is at some level a mystery, but to get
a better understanding we look at the predictions of some
basic network formation models.
We present a heuristic calculation based on the preferential attachment 
model, suggesting that a user's closure ratio should be related to the 
sum of the in-degrees of the user's followers, and 
we find on the Twitter data that the closure ratio indeed follows
this quantity more closely than simpler quantities such as the
user's own number of followers.
However, preferential attachment is not able to explain either 
the diversity of different closure ratios, or the fact that they
can be large on nodes of small in-degree; 
to understand these effects better, we analyze more complex models
that do not incorporate copying
as an overt or explicit mechanism in link formation,
including preferential attachment with fitness
\cite{bianconi-fitness} and a version of preferential
attachment with embedded community structure which is related to a
model of Menczer \shortcite{menczer-hier-links}.

We also note that the closure ratio of a user is distinct from --- and
exhibits qualitatively different properties than --- the 
{\em clustering coefficient} \cite{watts-strogatz}. 
The clustering coefficient is the fraction of pairs in a node's
neighborhood that are directly linked, and in the neighborhood of
a high-degree node it is almost always a small quantity,
for the fundamental reason that most of a high-degree node's neighbors
don't have enough incident edges to produce a significant
clustering coefficient \cite{vazquez-local-rules}.
The closure ratio, on the other hand, is a quantity that can
be quite large even for the neighborhoods of nodes with extremely
large degrees.

\begin{figure}[htp]
  \begin{center}
    \subfigure{\includegraphics[scale=0.4]{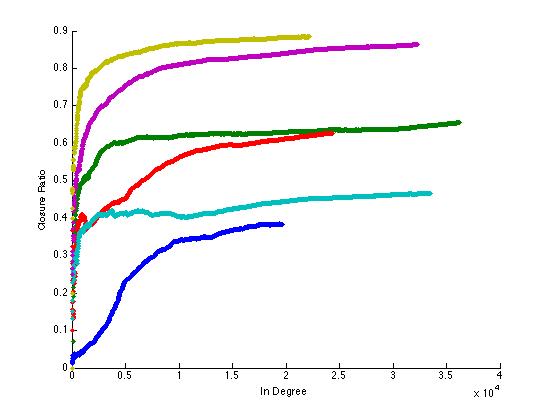}}
    \subfigure{\includegraphics[scale=0.4]{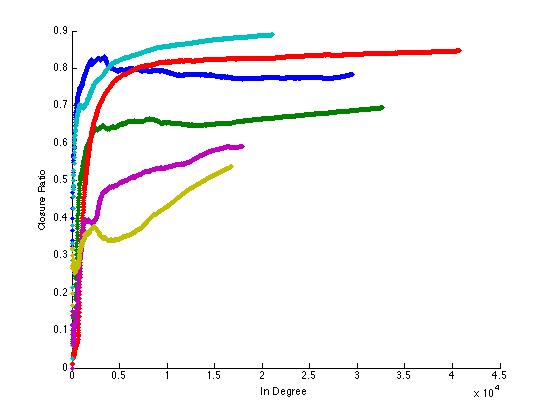}} \\
    \subfigure{\includegraphics[scale=0.4]{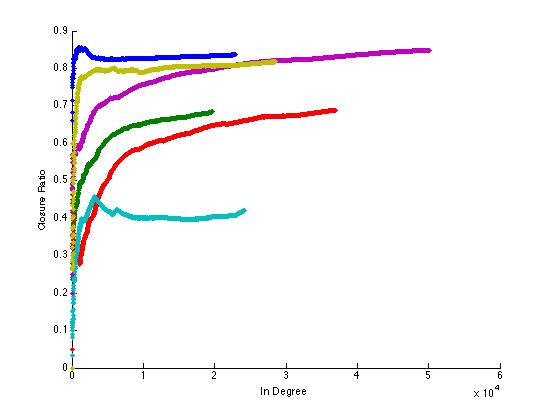}}
  \end{center}
  \caption{Closure ratio as a function of the arrival order of incoming edges for 18 Twitter $\mu$-celebrities. The following are the professions of the $\mu$-celebrities in each figure (from top to bottom curve). Top figure: Journalist, Venture Capital Blogger, Actor, Actor, DJ, Skateboarder. Middle figure: Comedian, Film Producer, Social Media Blogger, Musician, Actor, Journalist. Bottom figure: Comedian, TV Presenter, Actor, Musician, Filmmaker, Actor.}
  \label{trajectories}
\end{figure}

\section{Twitter Data and Micro-Celebrities}

We collected a random sample of $\mu$-celebrities on Twitter,
each with between 10,000 and 50,000 followers.
For each of these $\mu$-celebrities $C$, we determine the subset of
edges to $C$ that exhibit closure.

It is an interesting fact that determining this subset does not
require exact time-stamps or full network structure.
Rather, it is enough to have a chronologically ordered list
$L_{in}(C)$ of the followers of $C$, and for
each user $A \in L_{in}(C)$, a chronologically ordered list
$L_{out}(A)$ of the users that $A$ 
follows.\footnote{Such ordered lists were available via the Twitter API
at the time we performed these analyses
\cite{kalucki-twitter-chronological-url}.}
From these lists, we can conclude that an edge from $A$ to $C$
exhibits closure if and only if there exists a $B$ such that
$B$ precedes $A$ in $L_{in}(C)$ and $B$ precedes $C$ in $L_{out}(A)$.

In Figure~\ref{trajectories}, we show the running fraction of
edges that exhibit closure as the followers 
of a $\mu$-celebrity $C$ arrive in chronological order.
As noted in the introduction, in most cases this fraction reaches
a relatively stable value quite quickly, and this stable value
varies a lot from one $\mu$-celebrity to another.
Our models in the subsequent sections will help us investigate
this phenomenon.

\section{Evidence for Directed Closure}

We now use the randomization test described in the introduction to
identify evidence for the directed closure process at work.
We take the subgraph induced on the nodes in $\{C\} \cup L_{in}(C)$,
and we insert the edges in an order selected uniformly at random
from among all permutations of the edges.

Specifically,
we say that a user $A$ is {\em k-linked} to a user $C$ if 
$A$ follows $C$, and $A$ also follows $k$ followers of $C$.
(For example, in Figure~\ref{fig:multi-closure}, 
$A$ is 3-linked to $C$.)
Let $S_k(C)$ denote the set of all users who are $k$-linked to $C$,
and let $f_k$ denote the fraction of users in $S_k(C)$ whose
edge to $C$ exhibits closure.

Now, for each $k$ with $|S_k| > 10$,
we approximate the expected value of $f_k$ under the assumption that
the order in which the edges are created is chosen uniformly at random.
To do this, we run a simulation in which we 
generate a network consisting simply 
of a node $A$ pointing to a node $C$ and to $k$ other nodes
which also point to $C$; we randomly choose $|S_k|$ different orderings
of the edges of this network (one corresponding to each of the
$|S_k|$ followers who are $k$-linked to the real $\mu$-celebrity);
and we then determine the fraction of these random orderings
in which the $A$-$C$ edge exhibit closure.
We approximate the
expected value of $f_k$ over randomly ordered edges
by the average closure ratio among 100 runs of this 
simulation, and we define error bars using the minimum and the
maximum fraction among the 100 simulations. 

\begin{figure}[htp]
  \begin{center}
    \subfigure{\includegraphics[scale=0.4]{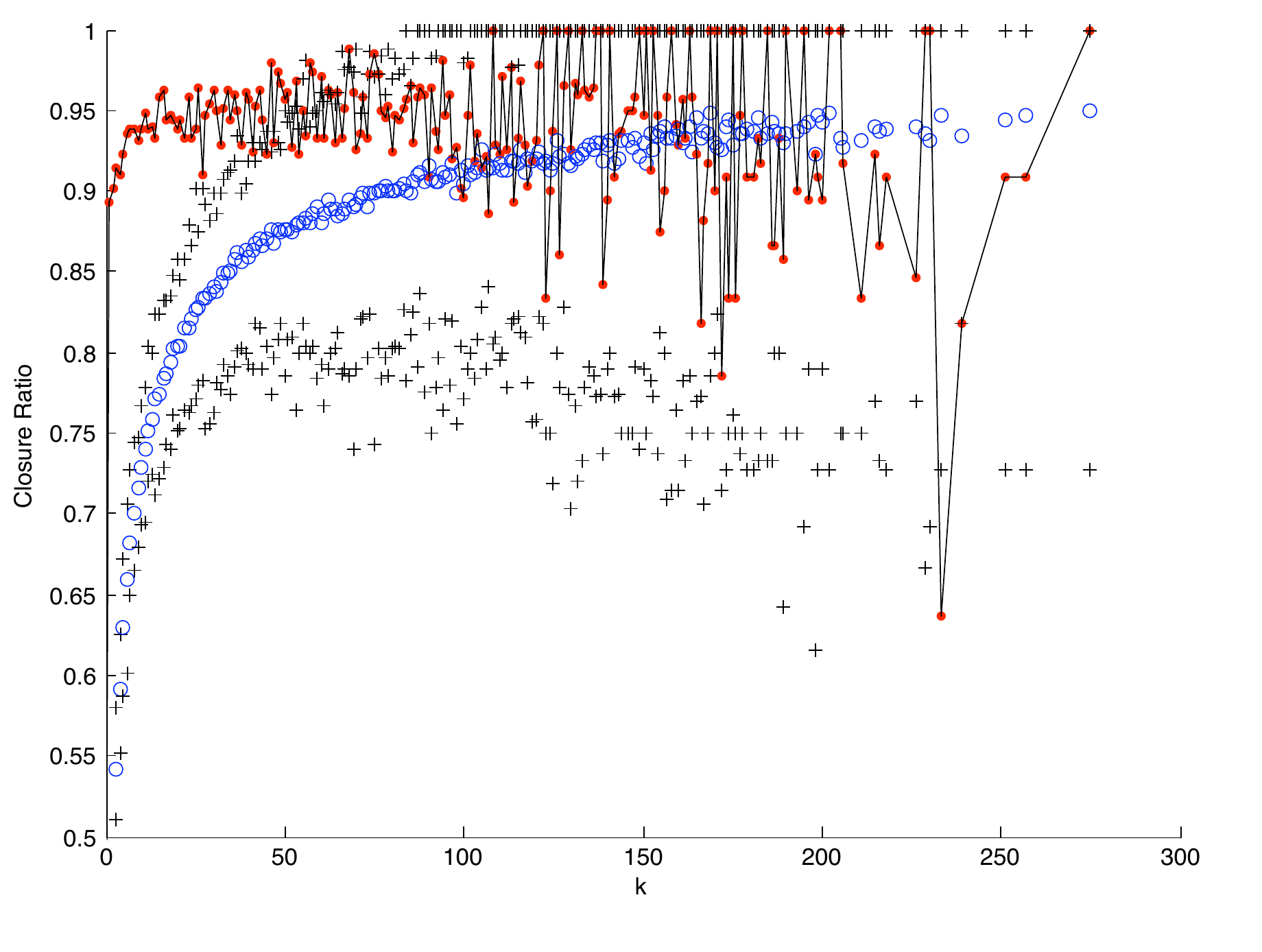}}
    \subfigure{\includegraphics[scale=0.4]{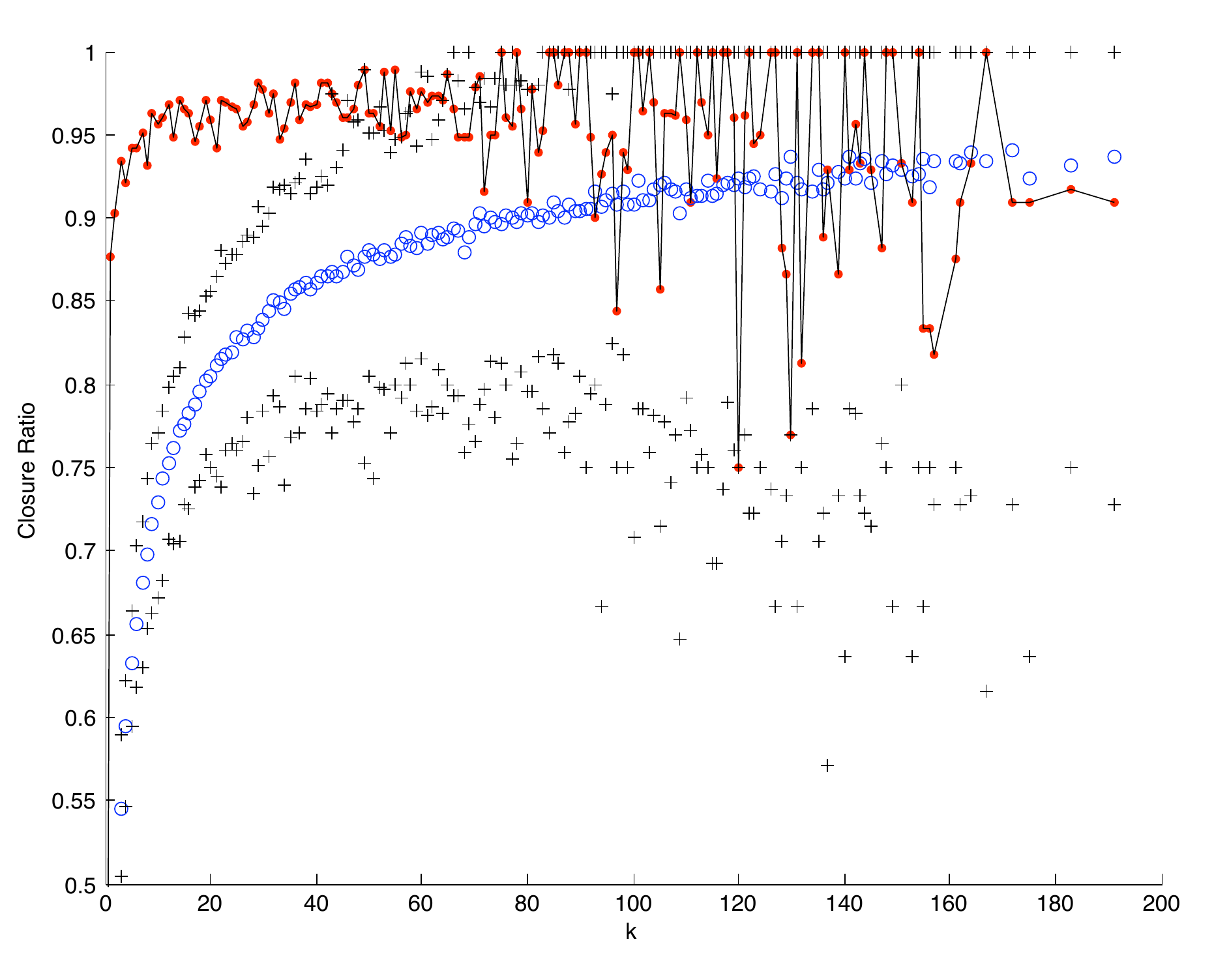}} \\
    \subfigure{\includegraphics[scale=0.4]{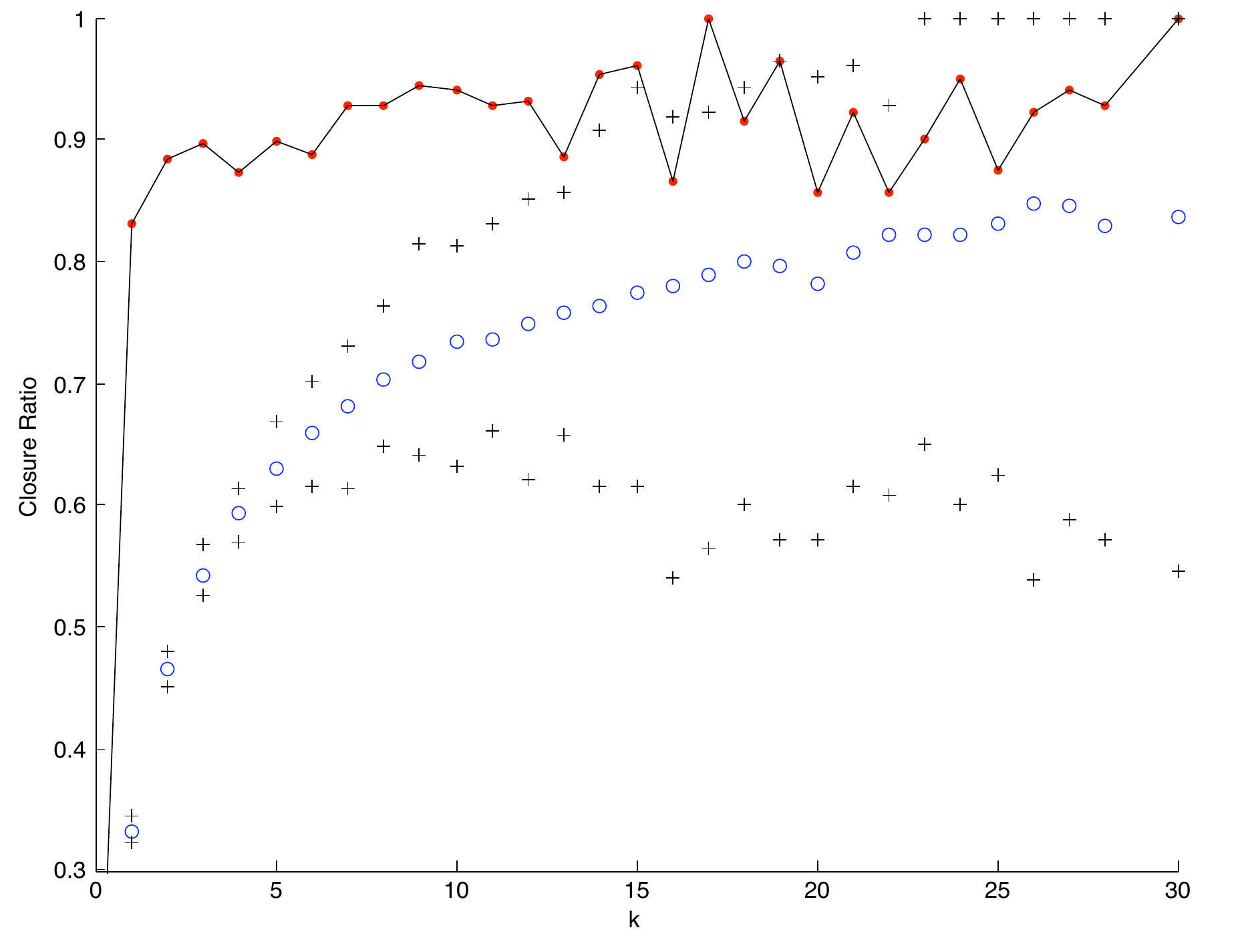}}
  \end{center}
  \caption{The connected dots indicate the actual value of $f_k$, the circles indicate the average closure ratio among the 100 simulations, and the plus signs indicate the error bars. Results for 3 $\mu$-celebrities are shown. The trend is similar for all other $\mu$-celebrities}
  \label{evid}
\end{figure}

We find the same trend for all the $\mu$-celebrities in our sample,
as shown in Figure~\ref{evid}: there
is some $K$ such that for all $k < K$ the actual value of $f_k$ is
higher than the maximum fraction from the 100 simulations. This means
that at least for small values of $k$ the fraction of edges exhibiting closure
is much higher than expected by chance.
This suggests the existence of an underlying mechanism --- copying
of links or something producing similar observed behavior ---
that makes it more likely than chance to see edges that appear to be copied.
For large values of
$k$, the expected value of $f_k$ assuming random ordering of edges becomes
very large, and it is hard for the values observed in the data to lie above
the error bars;
we find that for large $k$, the actual value of $f_k$ is very close to
the average fraction among the 100 simulations and is inside the
error bars.  

\section{Preferential attachment}

We would like to use probabilistic models of network formation 
to investigate the following
two fundamental properties of directed closure in the data.
First, for nodes whose in-degrees are at the level of $\mu$-celebrities,
the closure ratio saturates to a constant $f$ as edges arrive over time.
Second, this constant $f$ is quite different for
different $\mu$-celebrities, and it is not closely related to the total
in-degree of the $\mu$-celebrity. 

We now compare this with the predictions of a sequence of increasingly
complex models.
We begin with a very basic model --- a variant of the standard
\textit{preferential attachment} process, defined as follows
\cite{albert-revmodphys,newman-sirev}:
\begin{itemize}
\item Fix $\alpha \in [0,1]$, and $D,N \in \mathbb{N}$.
The graph will have $N$ nodes labeled ${0,1,2,...,N-1}$. 
\item Initially (at $t=0$) the graph 
consists of node labeled 1 with an edge 
pointing to the node labeled 0. 
\item At each time step ($t=j$) node $j$ will join the graph with $D$ edges
directed to nodes chosen from a distribution on 
$1,2,...j-1$. The endpoint of each edge is chosen in
the following way: With probability $\alpha$ the endpoint is chosen
uniformly at random from $\{1,2,...,j-1\}$. With probability
$1-\alpha$ the endpoint is chosen at random from a probability
distribution which weights nodes by their current in-degree.
\end{itemize}

\begin{figure}[tp]
  \begin{center}
   {\includegraphics[scale=0.4]{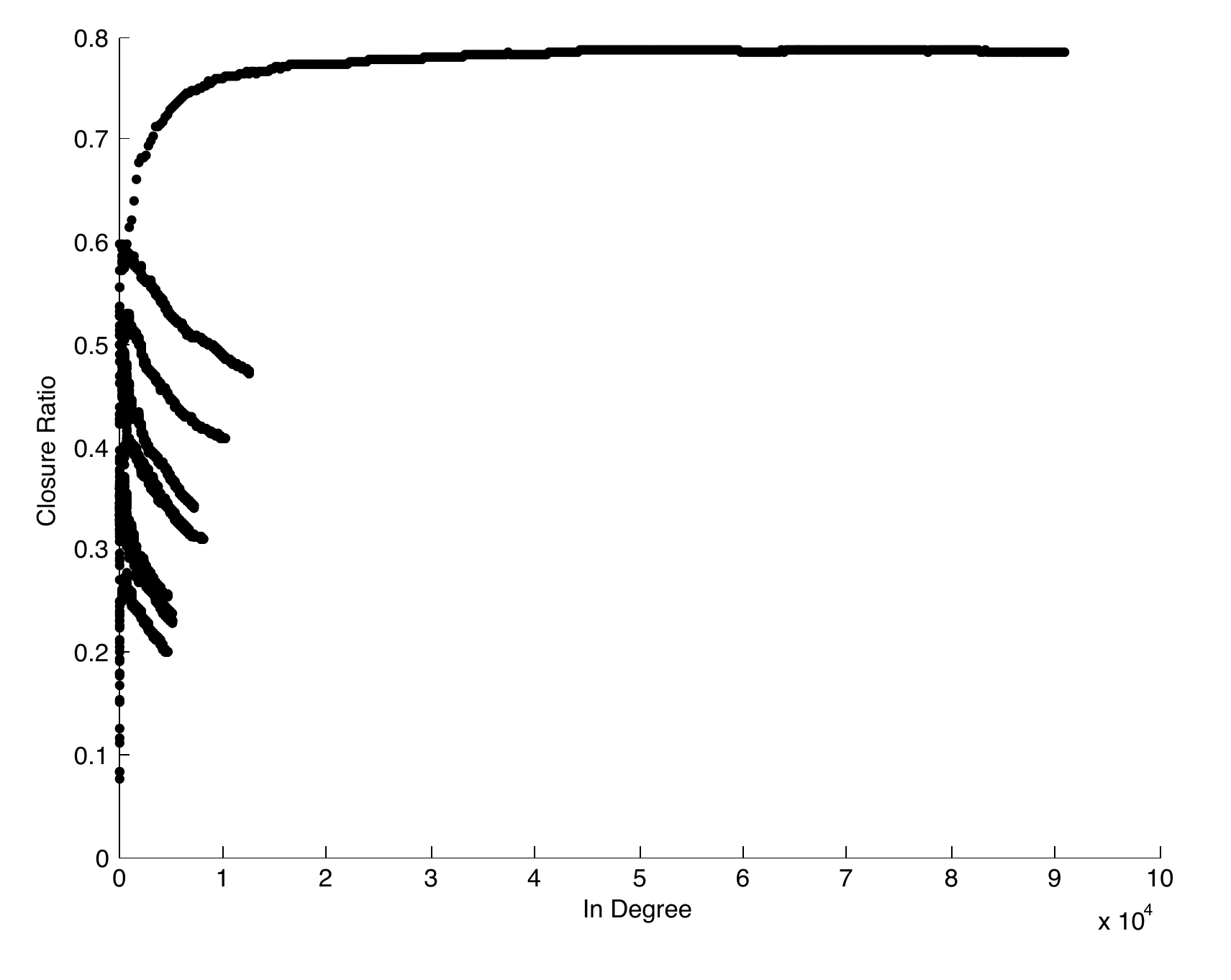}}
  \end{center}
  \caption{Results from the preferential attachment simulation with $N
  = 200,000$, $\alpha = .3$, and $D = 10$. The figure shows the 
  closure ratio
  as a function of edge arrival order of the 10 nodes with highest in-degree. }
  \label{prefAtt-saturation}
\end{figure}

We run this process with different values of $\alpha$, $D$, and $N$ and
find that preferential attachment does not achieve the desired results
for $\mu$-celebrities. In our simulations, only nodes with very large
in-degree have a reasonably large closure ratio,
while for other nodes it is essentially zero. For those nodes with very
large in-degree, the closure ratio saturates to a
constant $f$ as edges arrive, and the value of $f$ is different
for different nodes. However, the value of $f$ is monotonically
increasing as the final in-degree of node increases (See Figure
\ref{prefAtt-saturation}).

Through a heuristic calculation we now estimate the expected closure
fraction of a node in a graph generated by the
preferential attachment process.

Let 
$E_t $ be the total number of edges at time $t$, 
$N_t $ be the total number of nodes at time $t$,
$d_t(j)$ be the  in-degree of node $j$ at time $t$, 
$$F_t(j) = \{x : \exists e=(x,j) ~ \mbox{at time} ~ t\},$$ 
$d_t(S) = \displaystyle \sum_{x \in S} d_t(x)$, and 
$$S_t(j) = \alpha\frac{|F_t(j)|}{N_t} + (1-\alpha)  \frac{d_t(F_t(j))}{E_t}.$$
Note that $S_t(j)$ is the probability that a particular edge from node
$t+1$ is directed to a node $k$ such that there is an edge from $k$ to
$j$. In other words it is the probability that an edge from node $t+1$
is directed to a node that points to $j$.

Fix a node $j$ and an edge $e$ coming out of node $t+1$. We would like
to calculate the probability of the following event $V$: There is
another edge $e' = (t+1,x)$ created before $e$ such that $x$ points to
$j$ (i.e $\exists$ edge $g = (x,j)$). We will use $C_{t,e}(j)$ 
to denote the probability of this event $V$. 
Note that we do not know which of the $D$
edges coming out of $t+1$ the edge $e$ is, or what the destination of $e$ is.
Note that if $e$ is the first edge coming out of $t+1$ then
the event $V$ cannot happen;
if $e$ is the second edge coming out of
$t+1$ then $C_{t,e}(j) = S_t(j)$, if $e$ is the third edge coming out
of $t+1$ then $C_{t,e}(j) = [1-(1-S_t(j))^2]$, and more
generally if $e$ is the $d^{th}$
edge coming out of $t+1$ then $C_{t,e}(j) = [1-(1-S_t(j))^{d-1}]$.
Since it is equally likely that $e$ is any of the $D$ edges coming out
of $t+1$ we write
\begin{eqnarray*}
C_{t,e}(j) &=& \frac{1}{D}[1-(1-S_t(j))] 
   + \frac{1}{D}[1-(1-S_t(j))^2]+  \\ & &
\dots +\frac{1}{D}[1-(1-S_t(j))^{D-1}] \\
 &=& 1-\frac{1-(1-S_t(j))^D}{DS_t(j)}.
\end{eqnarray*}

If we knew that edge $e$ pointed to node $j$ then the event $V$ exactly says that $e$ exhibits closure. Therefore if we want to know the probability that $e$ exhibits closure given that $e=(t+1,j)$ we would need to calculate $P(V|e=(t+1,j))$. 
For the sake of our approximation, we use the unconditional
probability $P(V) = C_{t,e}(j)$ instead as our estimate of the probability
that $e$ exhibits closure. Note that the quantity $C_{t,e}(j)$ only
depends on $j$ and $t$, so we define
$C_t(j)=1-\frac{1-(1-S_t(j))^D}{DS_t(j)}$.  In general, a given edge
$e=(x,y)$ exhibits closure with a probability of approximately
$C_{x-1}(y)$. If $\displaystyle \lim_{t \rightarrow \infty} C_t(j) = L
< \infty$ then, for a large enough $T$, if $t > T$ then $C_t(j)
\approx L$. In other words, if $t > T$ the probability that an edge
coming out of node $t$ directed to node $j$ exhibits closure is
approximately $L$, which in turn is approximately $C_t(j)$. Therefore,
if $\displaystyle \lim_{t \rightarrow \infty} C_t(j) = L < \infty$ and
our parameter $N$ is large enough then $C_t(j) \approx C_{N-1}(j)$ for
$t > T$. Hence, if $N$ is large enough the final closure ratio of node
$j$ is approximately $C_{N-1}(j)$. 

\begin{figure}[tp]
  \begin{center}
    {\includegraphics[scale=0.4]{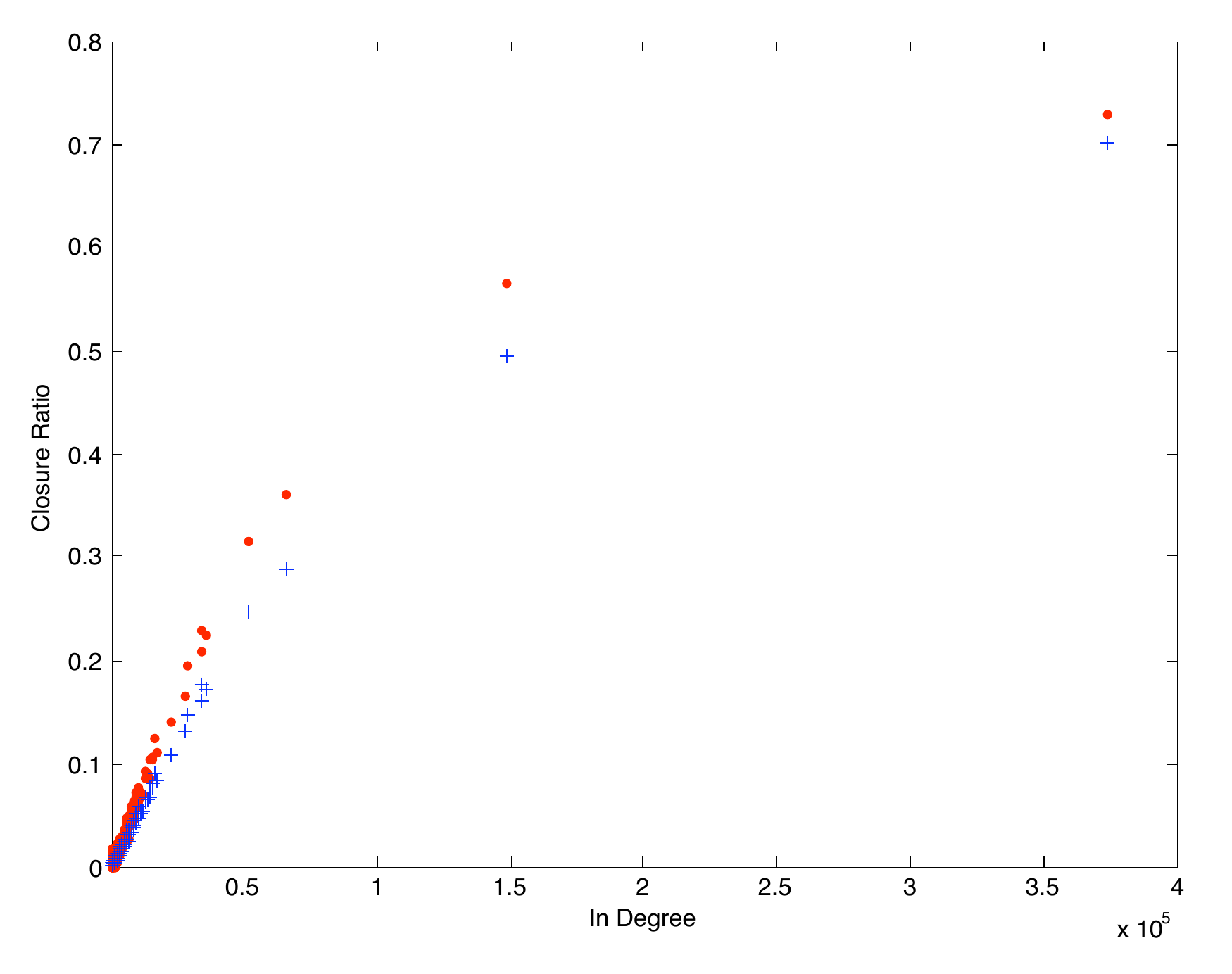}}
  \end{center}
  \caption{The actual closure ratio of each node $j$ generated by the preferential attachment model with parameters $N
  = 200,000$, $\alpha = .3$, and $D = 10$ (dots) and its approximation by $C_{N-1}(j)$ (plus signs). } 
  \label{prefAtt-closure-ratio}
\end{figure}

In Figure \ref{prefAtt-closure-ratio} we show that despite the 
approximations made in this argument, the calculation is 
a close fit to the actual closure ratios.

\begin{figure}[htp]
  \begin{center}
   \subfigure{\includegraphics[scale=0.4]{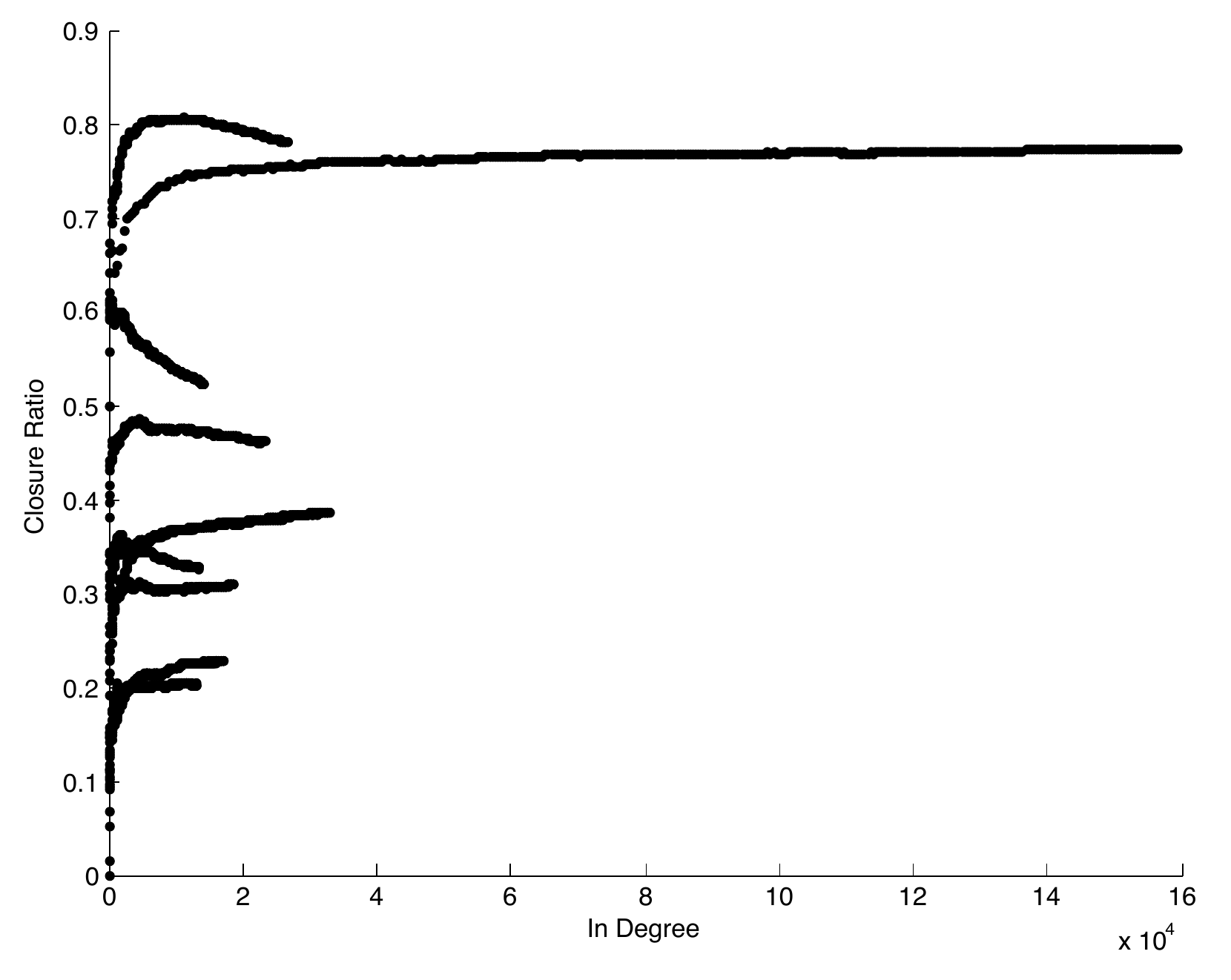}}\\
    \subfigure{\includegraphics[scale=0.4]{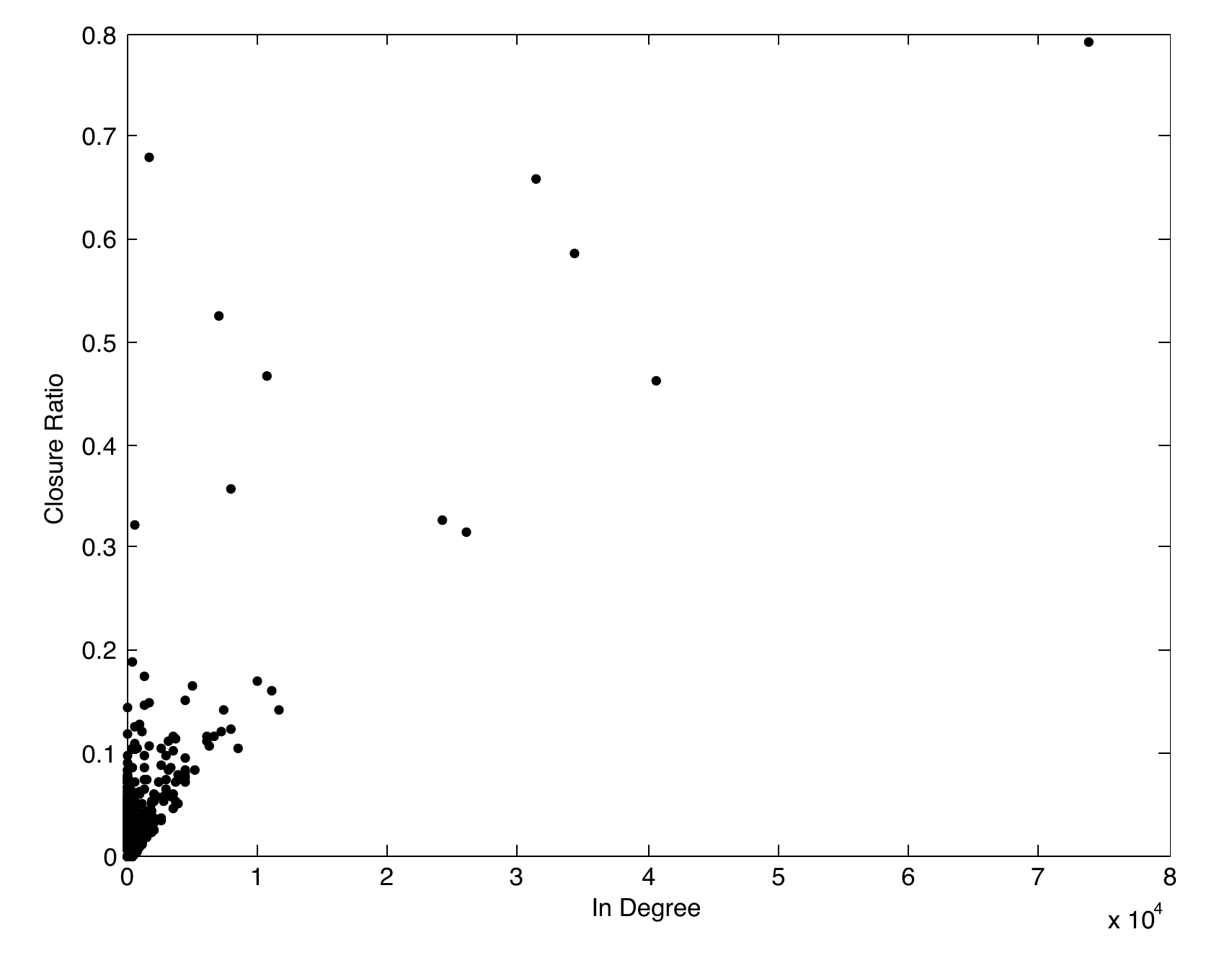}}
    \subfigure{\includegraphics[scale=0.4]{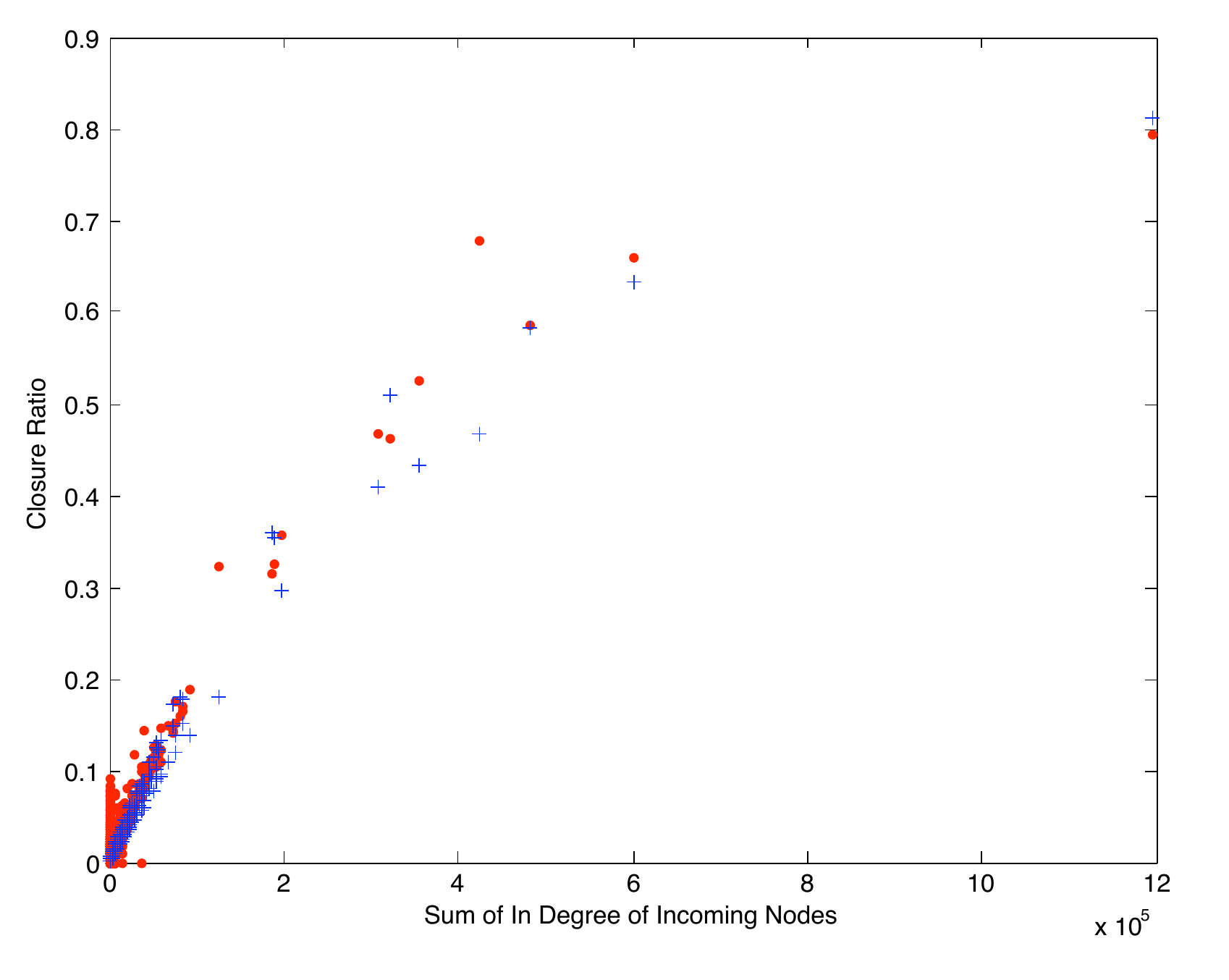}}
     
  \end{center}
  \caption{Results from the preferential attachment with fitness simulation with $N = 200,000$, $\alpha = .3$, and $D = 10$. The top figure shows the closure ratio as a function of in-degree of the 10 nodes with highest in-degree. The bottom function shows the final closure ratio of each node $j$ (dots) and its approximation by $C_{N-1}(j)$ (plus signs). }
  \label{prefAttFitness}
\end{figure}

\section{Preferential Attachment with Fitness}

The fact that preferential attachment produces very few nodes
with non-trivial closure ratios, and that these closure ratios
are closely tied to the in-degrees, indicates the need for a
more complex model.
One alternative would be the use of 
{\em copying models} \cite{kumar-copying,vazquez-copying},
where nodes explicitly copy links from other nodes that
have already joined the network.
Such a mechanism builds copying into the model, generally with a tunable
parameter that could be used to control quantities such as the closure ratio.
However, we would like to understand whether non-trivial closure ratios ---
and in particular, high levels of diversity in closure ratios --- 
can also appear in networks arising from models that do not
explicitly define copying as a mechanism.
As a first step in this direction, we investigate an extension of 
preferential attachment incorporating
the idea that different nodes may have different levels of
inherent {\em fitness} or {\em attractiveness}, which affects
how strongly they attract links 
\cite{bianconi-fitness}.

Here is how this model works:
\begin{itemize}
\item Fix $\alpha \in [0,1]$, and $D,N \in \mathbb{N}$.
The graph will have $N$ nodes labeled ${0,1,2,...,N-1}$. 
\item Each node also has a fitness
parameter $f_i \in(0,1)$ chosen uniformly at random. 
\item Initially (at $t=0$) the graph 
consists of node labeled 1 with an edge 
pointing to the node labeled 0. 
\item At each time step ($t=j$) node $j$ will join the graph with $D$ edges
directed to nodes chosen from a distribution on 
$1,2,...j-1$. The endpoint of each edge is chosen in
the following way: With probability $\alpha$ the endpoint is chosen
uniformly at random from $\{1,2,...,j-1\}$. With probability
$1-\alpha$ the endpoint is chosen at random from a probability
distribution which weights each node $i$ by $d_if_i$, where $d_i$ is
the node's current in-degree.
\end{itemize}

We run simulations of preferential attachment with fitness,
with different parameters, and find an improvement from the simple
preferential attachment model. A node's final closure ratio is 
not correlated with the final in-degree of the node,
which matches what we found in our data set. However, just like in the
simple preferential attachment model, very few nodes have a closure
fraction that is non-trivially larger than $0$
(see Figure \ref{prefAttFitness}).  
In particular, for the nodes that would
correspond to $\mu$-celebrities, the fraction is basically zero. This
is not consistent with the data, which shows that $\mu$-celebrities can
have very large closure ratios.

\begin{figure}[t]
  \begin{center}
   \subfigure{\includegraphics[scale=0.4]{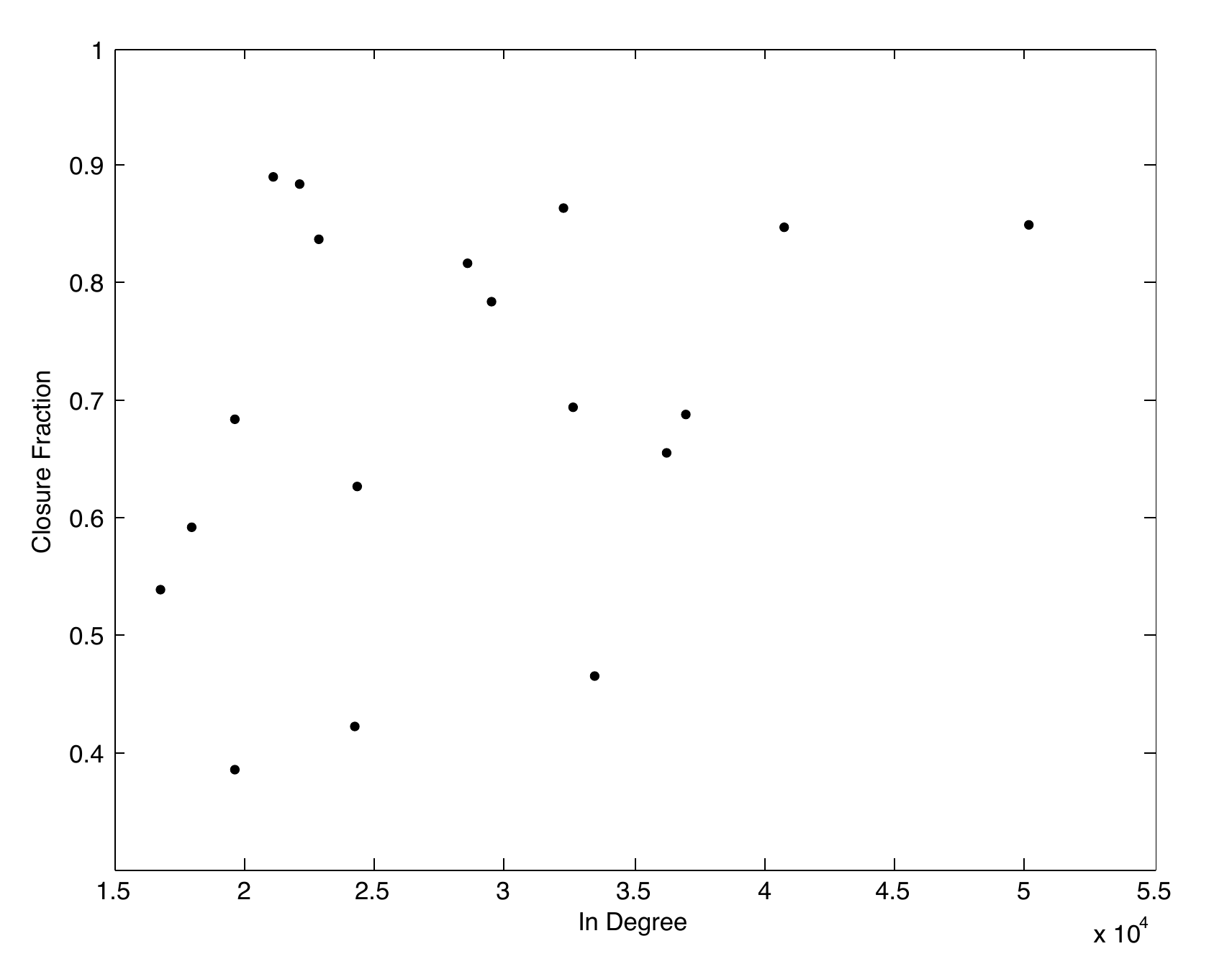}}
  \end{center}
  \caption{Closure ratio as a function of In-Degree.}
  \label{DataInDegVsFrac-indeg}
\end{figure}

\begin{figure}[t]
  \begin{center}
    {\includegraphics[scale=0.4]{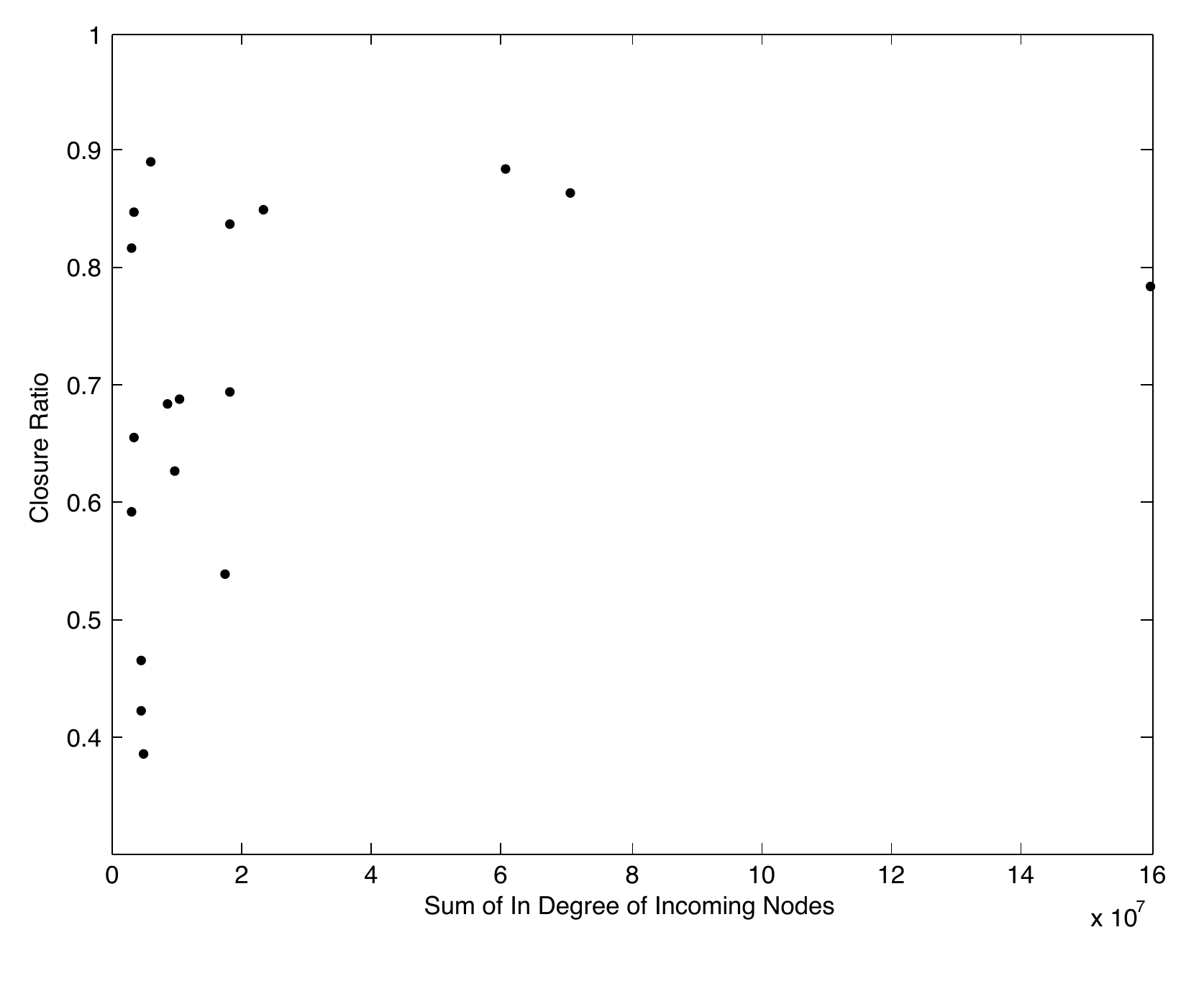}}
  \end{center}
  \caption{Closure ratio as a function of the Sum of In-Degree of Incoming Nodes.}
  \label{DataInDegVsFrac-indeg-sum}
\end{figure}

We find that the heuristic calculation for the closure ratio 
we derived for the preferential attachment model is
very accurate for preferential attachment with fitness as well. 
Furthermore, from the calculation we see that for a node $j$ the
term $d_t(F_{N-1}(j))$ (the sum of the in-degree of nodes that point
to $j$) is the most important in determining the closure ratio 
when $\alpha$ is small.  For preferential
attachment with fitness, the closure ratio 
of a node $j$ is much more correlated with $d_t(F_{N-1}(j))$ than with
the in-degree of $j$ (see Figure \ref{prefAttFitness}). This is also
the case for the $\mu$-celebrities in our data set (see Figures
\ref{DataInDegVsFrac-indeg} and \ref{DataInDegVsFrac-indeg-sum}),
which means that in determining a user's
closure ratio, the more important variable seems to be 
not the number of followers the user has but the total number of followers
of those who follow the user.

\section{Preferential Attachment with Communities}

The previous model, incorporating fitness, manages to produce 
heterogeneity in the closure ratios, but it still only produces
very few nodes for which the closure ratios are non-trivial.
We now present a model in which many nodes will have 
non-trivial closure ratios.

The model is 
\textit{preferential attachment with communities}: we assume that
each node belongs to a particular community of nodes, and the node is 
more likely to attach to nodes from its own community than to nodes
from other communities.
Specifically:
\begin{itemize}
\item Fix  $\alpha \in [0,1], \beta \in [.5,1]$, and 
$C, D$, and $N \in \mathbb{N}$.
The graph will have $N$ nodes labeled ${0,1,2,...,N-1}$ and there will
be $C$ communities. 
\item Initially (at $t=0$) the graph consists of the $C$
communities, each with two nodes, one pointing at the other. 
\item At each
time step ($t=j$) node $j$ will join the graph and will be assigned a
community uniformly at random. Then $j$ will create $D$ edges 
directed to nodes chosen from a distribution on
$1,2,...j-1$. The endpoint of each edge is chosen in the
following way: With probability $\beta$ the endpoint will be a node
from the same community as $j$; with probability $1-\beta$ the endpoint
will be chosen from any of $1,2,...j-1$. With probability $\alpha$ the
endpoint will be chosen preferentially (i.e. at random from a
probability distribution which weights nodes by their current
in-degree) and with probability $1-\alpha$ the endpoint will be chosen
uniformly at random from the set of nodes already determined.
\end{itemize}

\begin{figure}[tp]
  \begin{center}
  \includegraphics[scale=0.4]{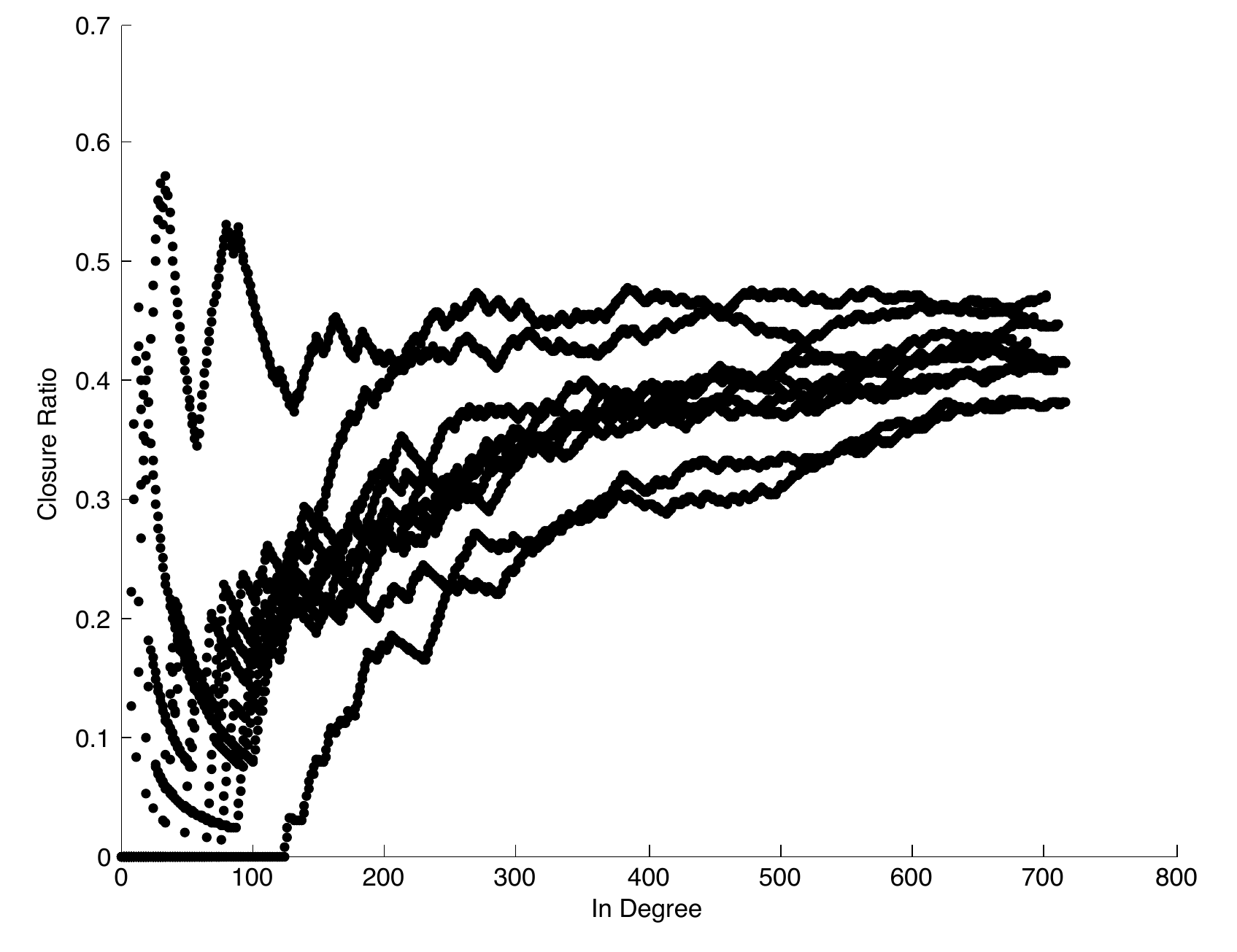}
  \end{center}
  \caption{The closure ratio as a function of in-degree 
for the 10 nodes with highest in-degree. 
Preferential attachment with communities simulation with $N = 200,000$, $\alpha = .3$, $\beta = .8$, $C = 1,000$, and $D = 10$. }
 \label{prefAttCommunityTrajects}
\end{figure}

Simulations with different parameters show that this model generates
nodes whose closure ratios converge as in-degree
increases (see Figure \ref{prefAttCommunityTrajects}), and the final
fraction is not closely related to the in-degree as it was in the
case of simple preferential attachment. Furthermore, the nodes that
would correspond to a $\mu$-celebrity level of in-degree 
can have reasonably large closure ratios.

\begin{figure}[htp]
  \begin{center}
  \subfigure{\includegraphics[scale=0.3]{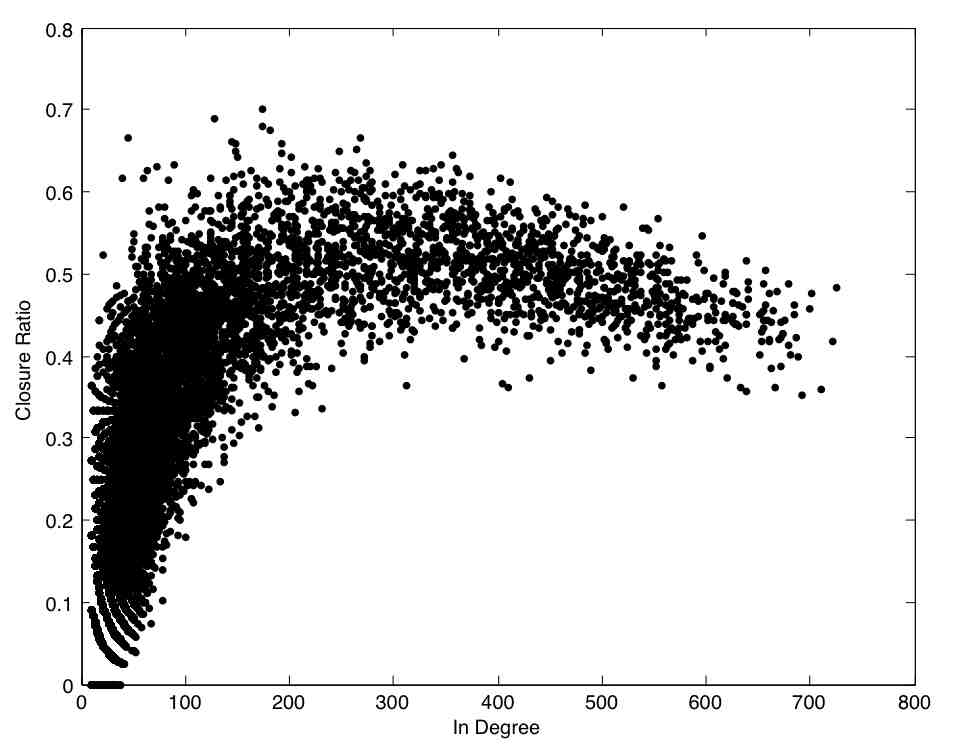}}
    \subfigure{\includegraphics[scale=0.3]{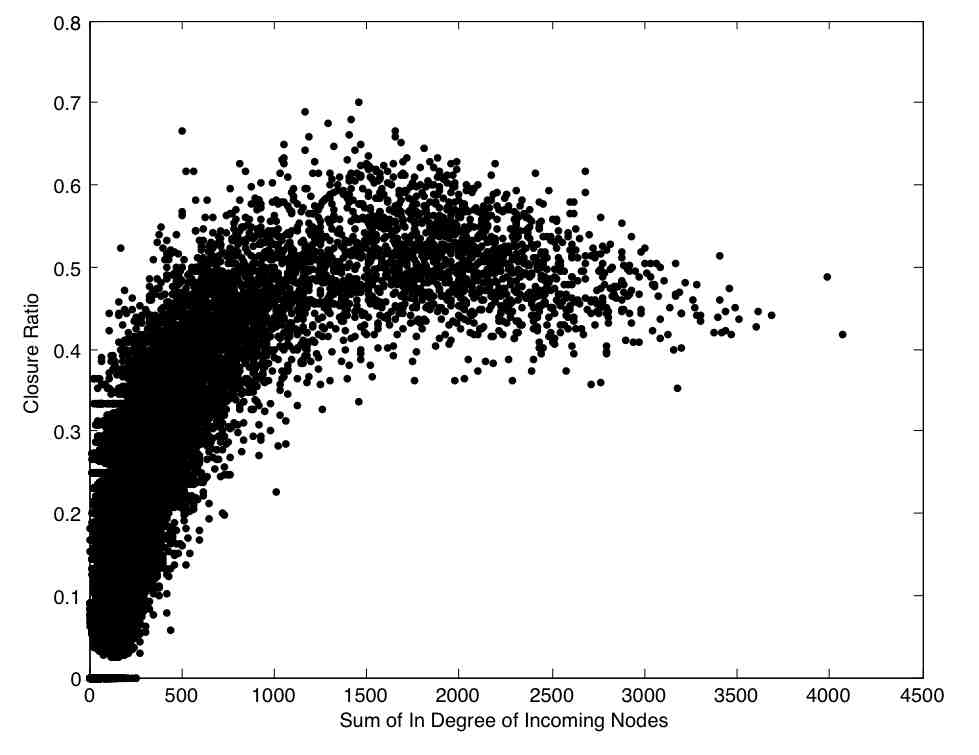}}
    \subfigure{\includegraphics[scale=0.3]{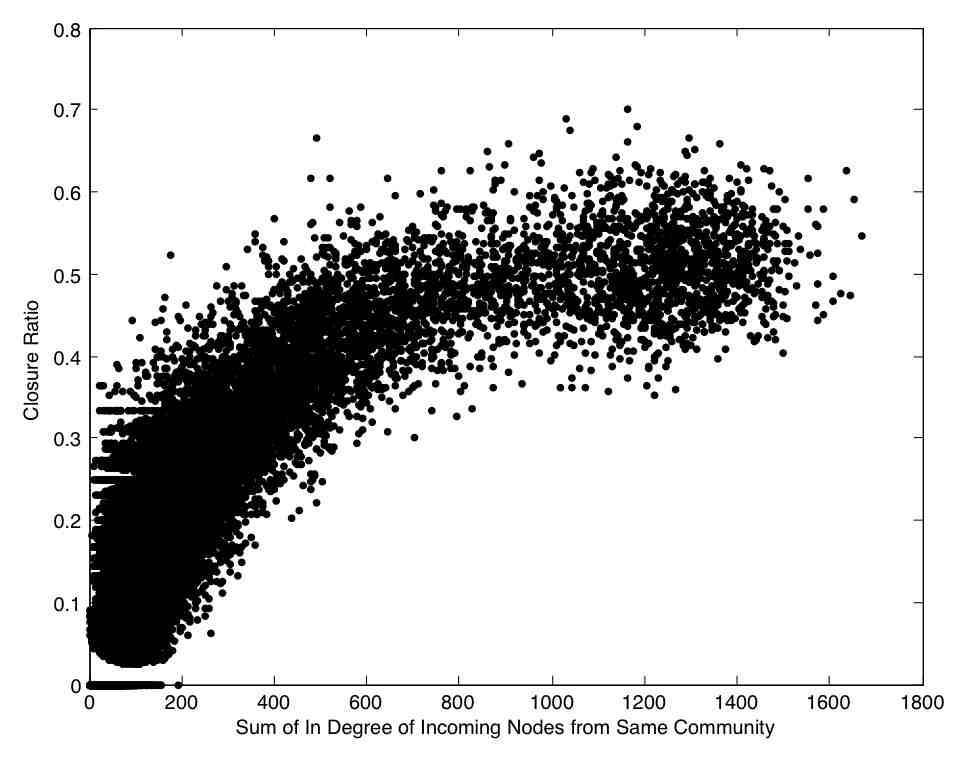}}
     
  \end{center}
  \caption{Results from the preferential attachment with communities simulation with $N = 200,000$, $\alpha = .3$, $\beta = .8$, $C = 1000$, and $D = 10$.}
  \label{prefAttCommunity}
\end{figure}

It is also interesting to note that the sum of a node's
followers' in-degrees, an important parameter in the previous
two models, still plays a role here, but with a twist:
as Figure~\ref{prefAttCommunity} shows, a node's closure 
ratio is more closely correlated with the sum of in-degrees
of the followers {\em from its own community} than with
the sum of the in-degrees of all its followers.
It would be interesting to explore this quantity on the
Twitter data, using different approximations of community structure 
in Twitter.

\section{Conclusion}

We have studied the process of directed closure in information networks,
developing a definition and methodology for evaluating it,
and providing evidence for directed closure in 
the follower network of Twitter.
We also found that the extent of directed closure varies considerably
between the sets of followers of different popular users.
A sequence of models generalizing the principle of preferential 
attachment provide some explanation for our findings, and identify
a more subtle parameter --- the sum of the in-degrees of one's followers ---
that is related to the extent of directed closure.

It is an interesting direction for further work to try understanding
better the causes of heterogeneity in the closure ratios of
micro-celebrities on Twitter, and the extent to which identifying
communities in the Twitter network structure can help evaluate
the more detailed predictions of preferential attachment with communities.
It will also be interesting to explore comparative analyses of 
these measures on other information networks.

\bibliography{n}
\bibliographystyle{aaai}
\end{document}